%% file: 2016-iser-popovic.tex
\documentclass[runningheads,a4paper]{llncs}

\usepackage{amssymb}
\usepackage{graphicx}
\usepackage{amsmath}
\usepackage{xspace}
\usepackage{algorithm}
\usepackage[noend]{algpseudocode}
\usepackage{color}
\usepackage{gensymb}
\usepackage{subcaption}
\usepackage[font=small]{caption}
\usepackage{wrapfig}
\usepackage[margin=1.7in]{geometry}

\usepackage{url}

\DeclareMathOperator*{\argmax}{\arg\!\max}

\graphicspath{{figures/}{generated-figures/}}

\usepackage{ifpdf}		
\ifpdf
	\usepackage[update]{epstopdf}
\else
\fi

\usepackage{hyperref}
\hypersetup{colorlinks,breaklinks,
linkcolor=[rgb]{0.5,0.,0.},
citecolor=[rgb]{0.000,0.427,0.173},
urlcolor=[rgb]{0.031,0.318,0.612}}

\usepackage{caption}

\usepackage[numbers,sectionbib,sort&compress]{natbib}

\usepackage{booktabs}

\usepackage[printonlyused,withpage,nolist,nohyperlinks]{acronym}
\input{common_acronyms.tex}

\providecommand{\figref}[1]{\mbox{Fig.~\ref{#1}}}
\providecommand{\eqnref}[1]{\mbox{Eq.~\ref{#1}}}
\renewcommand{\algref}[1]{\mbox{Alg.~\ref{#1}}}
\providecommand{\secref}[1]{\mbox{\textsection\ref{#1}}}
\title{Online Informative Path Planning \\for Active 
Classification on UAVs} % with Application to Precision Agriculture

\titlerunning{Informative Path Planning for Active Classification on UAVs}

\author{Marija Popovi\'{c} %\inst{1}
\and Gregory Hitz %\inst{1}
\and Juan Nieto %\inst{1}
\and Roland Siegwart %\inst{1} 
\and Enric Galceran %\inst{1}
}%
\authorrunning{Marija Popovi\'{c} et al.}
\tocauthor{
Marija Popovic (ETH Z\"{u}rich),
Gregory Hitz (ETH Z\"{u}rich),
Juan Nieto (ETH Z\"{u}rich),
Roland Y. Siegwart (ETH Z\"{u}rich),
Enric Galceran (ETH Z\"{u}rich)
}
\institute{
ETH Z\"{u}rich, Autonomous Systems Lab}%\\Leonhardstrasse 21, 8092 Z\"{u}rich, Switzerland.
\begin{document}

\mainmatter  % start of an individual contribution

\maketitle

\vspace{-6mm}

\begin{figure}[H]
\centering
\begin{subfigure}[]{0.065\textwidth}
  \raisebox{3.5mm}{\includegraphics[scale=0.52]{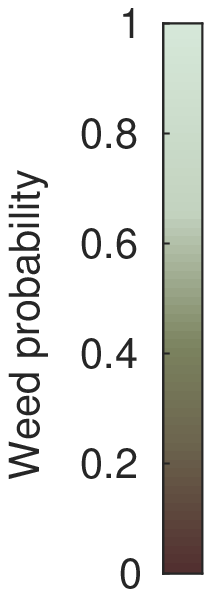}}
     \end{subfigure}
\hspace{2.5mm}
\begin{subfigure}[]{0.28\textwidth}
  \includegraphics[width=\textwidth]{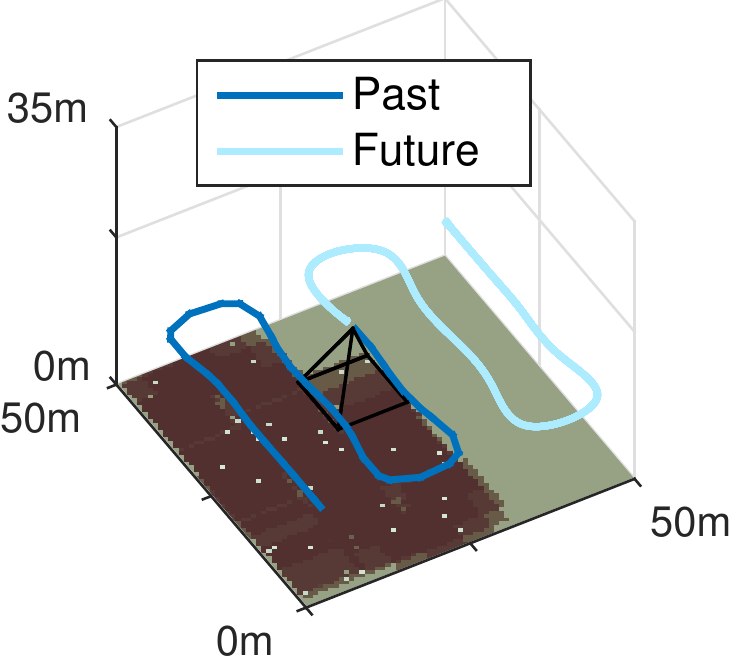} 
 % \caption{}
 % \label{SF:coverage_planner}
     \end{subfigure}
\begin{subfigure}[]{0.28\textwidth}
  \includegraphics[width=\textwidth]{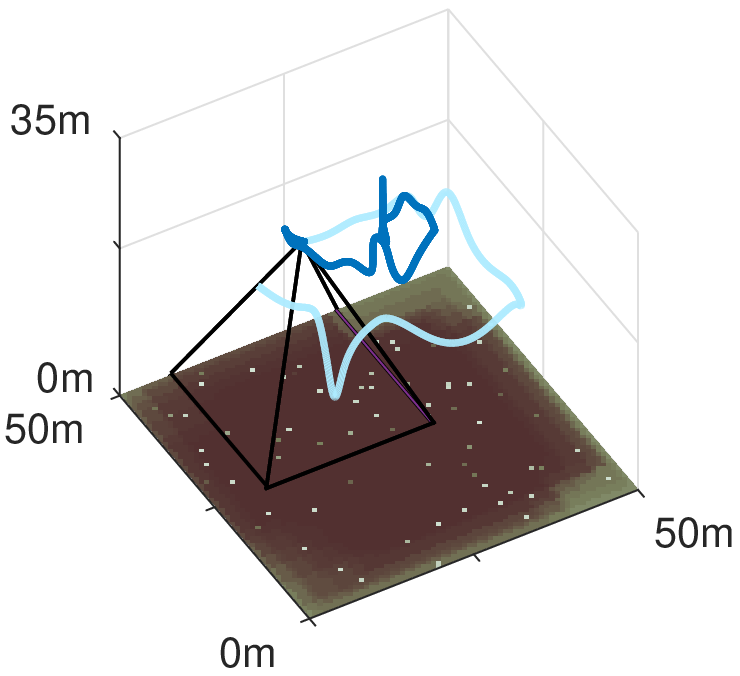}
 %   \caption{}
 %   \label{SF:our_planner}
     \end{subfigure}
\begin{subfigure}[]{0.28\textwidth}
  \includegraphics[width=\textwidth]{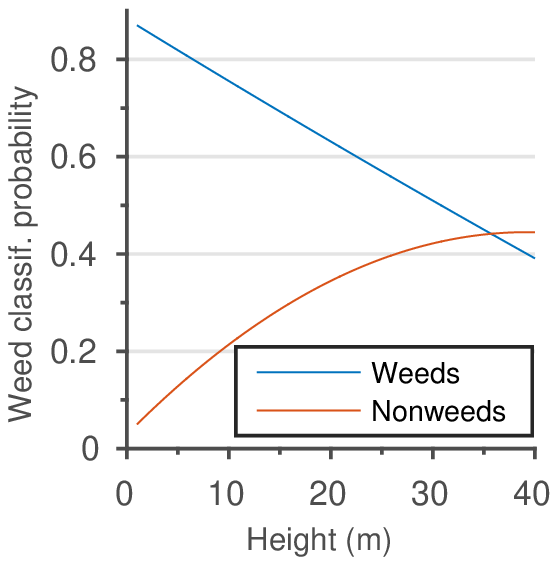}
 %   \caption{}
 %   \label{SF:meas_model}
     \end{subfigure}
\caption{By planning adaptively, our IPP approach (center) produces a weed map for precision agriculture 
with over half the entropy of a ``lawnmower'' coverage path (left) in the same time period. Our sensor 
model (right) allows better weed classifier performance with images taken at lower 
altitudes. The pyramid shows the camera 
footprint.} \label{F:teaser}
\end{figure}

\vspace{-10mm}

\begin{abstract}
We propose an \ac{IPP} algorithm for active classification using an \ac{UAV}, 
focusing on weed detection in precision agriculture. We model the 
presence of weeds on farmland using an occupancy grid and generate plans 
according to information-theoretic objectives, enabling the \ac{UAV} to gather 
data efficiently. We use a combination of 
global viewpoint selection and evolutionary optimization to refine the 
\ac{UAV}'s trajectory in continuous space while satisfying dynamic 
constraints. We validate our approach in simulation by comparing against 
standard ``lawnmower'' coverage, and study the effects of varying objectives 
and optimization strategies. We plan to evaluate our algorithm on a real 
platform in the immediate future.
\end{abstract}

\acresetall

\section{Introduction}

Autonomous robots are increasingly used to gather information about the 
Earth's ecosystems~\cite{Dunbabin2012a}. In agricultural monitoring, \acp{UAV} are capable 
of providing high-resolution data in a flexible, cost-efficient 
manner~\cite{Detweiler2015}. Using sensors, 
\acp{UAV} can survey crops to find precision treatment 
targets, improving yield and leading to sustainability and economic 
gain~\cite{Cardina1997}. Unfortunately, \acp{UAV} are often constrained by limited
battery and computational capacities. Therefore, planning for efficient data 
collection is key in enabling robotics in this field.

We address the problem by proposing an \ac{IPP} algorithm for active 
classification on a UAV equipped with an image-based weed classifier. We model 
the presence of weed on farmland using an occupancy grid. We continuously plan 
paths online through a combination of global viewpoint selection and 
evolutionary optimization, which refines the \ac{UAV}'s trajectory in 
continuous 3D space while satisfying dynamic constraints. The 
resulting informative paths abide by a limited time budget and
address the key challenge of trading off sensor resolution against coverage 
when flying at variable altitudes.

Our contributions are:
\begin{itemize}
 \item An IPP algorithm with the following properties:
 \begin{itemize}
  \item generates dynamically feasible trajectories in continuous space,
  \item obeys budget and sensing constraints,
  \item trades off sensor resolution against coverage in a principled manner by incorporating a 
height-dependent sensor noise model.
 \end{itemize}
 \item An evolutionary strategy to optimize continuous \ac{UAV} paths for maximum informativeness.
 \item Validation of our approach in simulation against a coverage planner.
\end{itemize}
We plan to evaluate our algorithm in field experiments in the immediate future.

\section{Related Work}

Most previous \ac{IPP} approaches seek to minimize map uncertainty using 
objectives derived from Shannon's entropy~\citep{Charrow2015a, Hollinger2014}. 
To exploit new data, adaptive approaches~\citep{Hitz2015, Lim2015, 
Girdhar2015} replan paths based on specific interests. IPP can be performed
using combinatorial optimization over a discrete 
grid~\citep{Chekuri2005,Hollinger2009,Binney2013}. However, the drawbacks of this 
representation are its limited scalability and resolution. Alternatively, some 
planners work in continuous space by leveraging sampling-based 
methods~\citep{Hollinger2014} or splines~\citep{Charrow2015a, 
Hitz2015, Marchant2014}. Similarly to~\citet{Charrow2015a}, we use global 
viewpoint selection to escape local minima and optimization to refine our 
trajectory.

IPP addressing UAV imaging is a relatively unexplored area. A set-up similar to 
ours has been studied recently 
by~\citet{Sadat2015}. However, their method assumes discrete viewpoints 
and prior knowledge of target regions, neglecting sensor noise. In 
contrast, our approach considers a height-dependent sensor model and 
incrementally replans as data are collected. Moreover, we use smooth polynomial 
trajectories which guarantee feasibility of the \ac{UAV}'s dynamic constraints.

\section{Problem Definition} \label{sec:problem_definition}

We define the general IPP problem as follows. We seek a continuous path $P$ in 
the space of all possible paths $\Psi$ for maximum gain in some information 
measure:
\begin{equation}
\begin{aligned}
  P^* ={}& \underset{P \in \Psi}{\argmax}
\frac{I[\textsc{measure}(P)]}{\textsc{time}(P)}\textit{,} \\
 & \text{s.t. } \textsc{time}(P) \leq B \textit{,}
 \label{E:ipp_problem}
\end{aligned}
\end{equation}
where $B$ denotes a time budget and $I$ quantifies the objective, 
discussed in~\secref{ssec:planning_algorithm} for our application. The 
function \textsc{measure(\textperiodcentered)} 
obtains measurements and \textsc{time(\textperiodcentered)} 
provides the travel time along the path. Maximizing information gain 
\textit{rate}, as opposed to maximizing only information, enables comparing the values of paths over 
different time scales.

\section{Technical Approach}
In this section, we present our IPP algorithm. The main idea is to 
create fixed-horizon plans maximizing an informative objective. To do this 
efficiently, we first select global viewpoints and then optimize the path in continuous space using an 
evolutionary method. In~\secref{ssec:modeling} and~\secref{ssec:path_parametrization}, we 
introduce our approaches to modeling and path 
parametrization. In~\secref{ssec:planning_algorithm}, we detail 
our planning routine, shown in~\algref{A:replan_path}.

\begin{algorithm}[h]
\renewcommand{\algorithmicrequire}{\textbf{Input:}}
\renewcommand{\algorithmicensure}{\textbf{Output:}}
\algrenewcommand\algorithmiccomment[2][\scriptsize]{{#1\hfill\(\triangleright\)
\textcolor[rgb]{0.4, 0.4, 0.4}{#2} }}
\begin{algorithmic}[1]

  \State $\mathcal{X}^g, \mathcal{X}^i \gets \emptyset$ \Comment{Initialize 
global and intermediate viewpoints.}
  \While {$\textit{H} \geq |\mathcal{X}^g \cup \mathcal{X}^i|$}
   \If {$\textit{t/B} < \Call{rand}$ } \Comment{Tradeoff global selection objectives based 
on time.}
   \State $\textbf{x}^* \gets$ Select viewpoint in $\mathcal{L}$ 
using~\eqnref{E:info_objective}
  \Else
    \State $\textbf{x}^* \gets$ Select viewpoint in $\mathcal{L}$ 
using~\eqnref{E:class_objective}
  \EndIf
  \State $\mathcal{M} \gets$ \Call{simulate\_measurement}{$\mathcal{M}$,
$\textbf{x}$} \Comment{Simulate using ML.}
  \State $\textit{t} \gets \textit{t}$ + \Call{time}{$\textbf{x}^*$}
  \State $\mathcal{X}^g \gets \mathcal{X}^g \cup \textbf{x}^*$; $\mathcal{X}^i \gets \mathcal{X}^i 
\cup
\Call{add\_intermediate\_points}{\textbf{x}^*}$
  \EndWhile
  \State $\mathcal{X} \gets \mathcal{X}^g \cup \mathcal{X}^i$; $\mathcal{X} \gets$ 
\Call{cmaes}{$\mathcal{X}$, $\mathcal{M}$}
\Comment{Optimize polynomial path.}

\end{algorithmic}
\caption{\textsc{replan\_path} procedure}\label{A:replan_path}
\end{algorithm}

\subsection{Environment and Measurement Models} \label{ssec:modeling}
We represent the environment (a farmland above which the \ac{UAV} flies) using 
a 2D occupancy grid map $\mathcal{M}$ \citep{Elfes1989}, where each cell 
is associated with a Bernoulli random variable representing the probability of 
weed occupancy. For our measurement model, we assume a square 
footprint for a down-looking camera providing input to a weed classifier. The 
classifier provides probabilistic weed occupancy for cells within \ac{FoV} 
from a UAV configuration $\textbf{x}$. For each observed cell 
$\textbf{m}_i \in \mathcal{M}$ at time $t$, we perform 
a log-likelihood update given an observation $z$:
\begin{equation}
  \text{logit}(p(\textbf{m}_i | z_{1:t}, \textbf{x}_{1:t})) = 
\text{logit}(p(\textbf{m}_i | z_{1:t-1}, \textbf{x}_{1:t-1})) + 
\text{logit}(p(\textbf{m}_i 
| z_t, \textbf{x}_t)) \text{,}
 \label{E:occupancy_grid_update}
\end{equation}
where the second term denotes the height-dependent 
sensor model capturing the weed classifier output. Our sensor model (\figref{F:teaser}, right) matches real 
datasets	 at low altitudes~\citep{Haug2014} and accounts for
poorer classification performance with lower-resolution 
images taken at higher altitudes.

\subsection{Path Parametrization} \label{ssec:path_parametrization}
To create paths abiding by the dynamic constraints of the \ac{UAV}, we connect 
viewpoints $\textbf{x}\in\mathcal{X}$ using the method of~\citet{Richter2013}. As 
in their work, we express a 12-degree polynomial trajectory in terms of 
end-point derivatives, allowing for efficient optimization in 
an unconstrained quadratic program.

\subsection{Planning Algorithm}  \label{ssec:planning_algorithm}
\begin{figure}[h]
\centering
\begin{subfigure}[]{0.255\textwidth}
  \includegraphics[width=\textwidth]{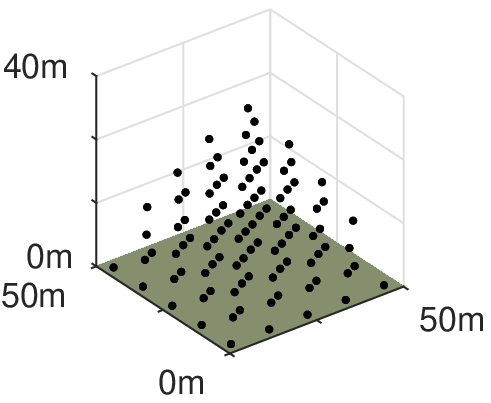} 
  \caption{}
  \label{SF:lattice}
     \end{subfigure}
\begin{subfigure}[]{0.195\textwidth}
  \includegraphics[width=\textwidth]{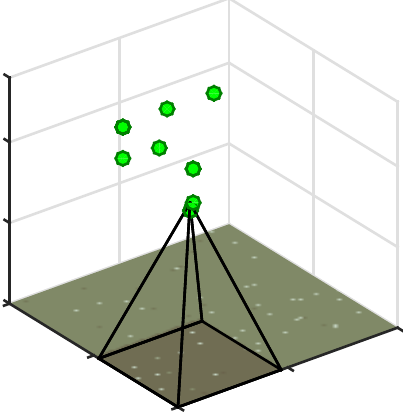} 
  \caption{}
  \label{SF:global_points}
     \end{subfigure}
\begin{subfigure}[]{0.195\textwidth}
  \includegraphics[width=\textwidth]{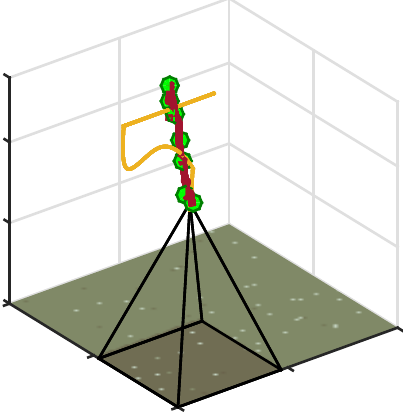}
    \caption{}
    \label{SF:global_optimization}
     \end{subfigure}
\begin{subfigure}[]{0.195\textwidth}
  \includegraphics[width=\textwidth]{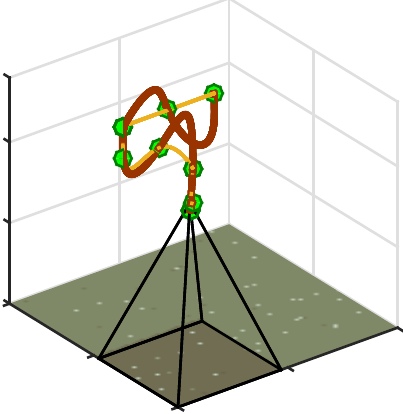}
    \caption{}
    \label{SF:intermediate_optimization}
     \end{subfigure}
\caption{Our planner uses a lattice (a) for selecting 
global viewpoints (b). The trajectory is then refined globally (c) or 
locally (d). The orange and maroon curves show paths before and after
optimization, respectively.}\label{F:replan_path}
\end{figure}

We use a fixed-horizon approach to plan adaptively. During the mission, we 
maintain viewpoints $\mathcal{X}$ within a horizon \textit{H}. We 
alternate plan execution and replanning, stopping when a time budget \textit{B} 
is exceeded. For replanning, we adopt a two-stage approach consisting of global 
viewpoint selection and optimization. This procedure is described in~\algref{A:replan_path} and illustrated 
in~\figref{F:replan_path}. 
The following sub-sections detail the key steps of~\algref{A:replan_path}.

\subsubsection{Global Viewpoint Selection} In the first step (Lines 
3-9), we sequentially select global measurement sites $\mathcal{X}^g$ 
(\figref{SF:global_points}). Unlike in frontier-based exploration, common for indoor 
mapping~\citep{Charrow2015a}, choosing viewpoints using map 
boundaries is not applicable in our set-up. Instead, we 
apply~\eqnref{E:ipp_problem} over the horizon \textit{H} (Line~2) to 
find most informative measurement sites. To find the next viewpoint 
$\textbf{x}^*$ efficiently, we evaluate the objective over a multiresolution 
lattice $\mathcal{L}$ (\figref{SF:lattice}). To encourage exploration, 
we maximize entropy reduction in $\mathcal{M}$:
\begin{equation}
  I  [t+1|t] = H(\mathcal{M}_{t}) - H(\mathcal{M}_{t+1}) \text{.}
 \label{E:info_objective}
\end{equation}
To encourage classification, we divide $\mathcal{M}$ into 
``weed'' and ``non-weed'' cells using thresholds $\delta_{w}$ and $\delta_{nw}$, leaving an unclassified
subset $\mathcal{U} = \{m_i \in \mathcal{M}\,|\,\delta_{nw} < p(m_i) < \delta_w\}$. This 
is similar to finding unknown space in conventional occupancy mapping. We maximize the reduction of 
$\mathcal{U}$ between time-steps:
\begin{equation}
  I[t+1|t] = |\mathcal{U}_{t}| - |\mathcal{U}_{t+1}| \text{.}
 \label{E:class_objective}
\end{equation}
We use an optional time-varying parameter (Line 3) to gradually bias viewpoint selection 
towards~\eqnref{E:class_objective} from~\eqnref{E:info_objective}, focusing on
weed identification over time. We then simulate a maximum likelihood (ML) 
measurement at $\textbf{x}^*$ (Line 7) and interpolate intermediate viewpoints 
$\mathcal{X}^i$ to add degrees of freedom to the path (Line 9).

\subsubsection{Optimization} In the second step (Line 10), we optimize the 
polynomial path by solving Eq. \ref{E:ipp_problem} in~\secref{sec:problem_definition} using the Covariance 
Matrix Adaptation Evolution Strategy (CMA-ES). CMA-ES is a gradient-free optimizer 
suitable for continuous shape fitting with our discrete measurement model 
\citep{Hansen2006}. As evaluated in~\secref{sec:results}, we consider (i) globally 
optimizing $\mathcal{X}$ (\figref{SF:global_optimization}) and (ii) optimizing 
$\mathcal{X}^i$ only (\figref{SF:intermediate_optimization}).

\section{Results} \label{sec:results}
Our algorithm is validated in simulation on 260 $50\times50$m environments with 
120 Poisson-distributed weeds. We use thresholds of $\delta_{nw} = 0.25$ and 
$\delta_{w} = 0.75$, and use a replanning horizon $H$ of 7 viewpoints to limit optimization complexity. We 
initialize the \ac{UAV} position as the 
map center with maximum altitude (45m). Our methods are evaluated against traditional ``lawnmower'' coverage 
with height fixed (8.66\,m) 
for the same 300\,s budget $B$. We consider map entropy, 
classification rate, and mean F1-score as metrics, common for classification tasks. We simulate sensor noise 
using our model in \secref{ssec:modeling}. Following a similar approach to~\citet{Pomerleau2013}, we 
compute the cumulative distribution function (CDF) of entropy over a time histogram to 
summarize variability among trajectories.

\pagebreak

In our experiments, we study varying:
\begin{itemize}
 \item Global viewpoint objectives: information only (\eqnref{E:info_objective}), classification only
(\eqnref{E:class_objective}), \\time-varying (\secref{ssec:planning_algorithm})
 \item Optimization methods: no CMA-ES, local CMA-ES, global CMA-ES
\end{itemize}

As described in \secref{ssec:planning_algorithm}, for local CMA-ES, we consider optimizing $\mathcal{X}^i$ to 
reduce inter-segment entropy. For global CMA-ES, we optimize $\mathcal{X}$, allowing points in 
$\mathcal{X}^g$ to vote on the optimization objective for the entire trajectory.

\begin{figure}[h]
\centering
\includegraphics[width=0.955\textwidth]{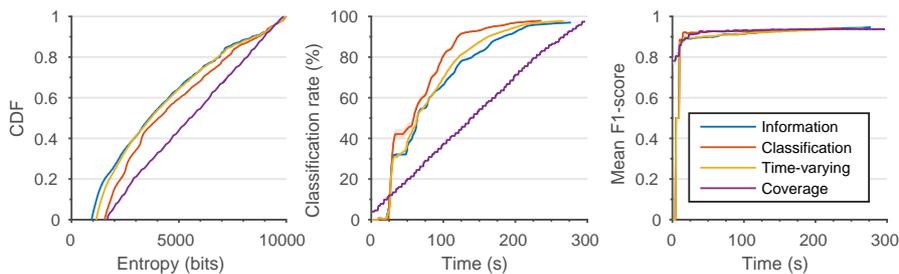}
\caption{Comparison of global viewpoint selection objectives (\secref{ssec:planning_algorithm}) with local 
CMA-ES optimization. The solid 
lines indicate means over 260 trials. The thin shaded regions depict 95\% 
confidence bounds. Using our methods, the metrics improve quickly as the UAV flies at variable altitudes. 
Accounting for 
spent budget (time) trades off the objectives.}
\label{F:objectives}
\end{figure}

\begin{figure}[!h]
\centering
\includegraphics[width=0.955\textwidth]{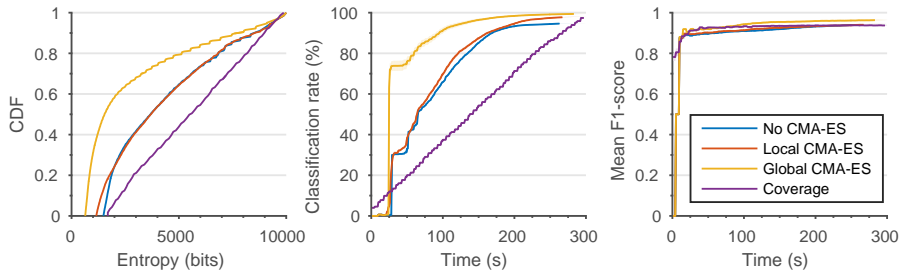}
\caption{Comparison of optimizers for the time-varying objective. 
The effect of local optimization is marginal due to the small number 
of refined points. Overall, global optimization performs best as the entire path can be varied.}
\label{F:optimizers}
\end{figure}

\figref{F:objectives} compares the global viewpoint selection objectives with local CMA-ES 
optimization. Our methods outperform na\"ive coverage as they permit variable-altitude flight 
for wider \acp{FoV}. As shown in~\figref{F:teaser}, our planners usually 
produce paths similar to spirals, starting with descent to the unknown map center. The curves 
illustrate the coverage-resolution trade-off: for the classification objective (red curve), flying at low 
altitudes quickly provides accurate classification, as shown by the rises in classification rate and 
F1-score. However, entropy reduction is limited. By considering elapsed time when selecting 
global viewpoints (yellow curve), we obtain high certainty with efficient classification. 
\figref{F:optimizers} compares the CMA-ES optimization methods for the time-varying objective. The global 
method likely performs best due to the 
highest number of 
optimized variables.


\section{Conclusion and Scheduled Experiments}
We presented an adaptive IPP strategy for active weed classification on UAVs. 
Our algorithm combines global viewpoint selection with evolutionary optimization 
to generate dynamically feasible paths with informative objectives. We validated 
our strategy against a ``lawnmower'' coverage pattern and demonstrated the effects of using 
different objectives and optimization strategies.

We aim to implement our algorithm on an AscTec Neo \ac{UAV} platform. Our experiments will take 
place at the ETH Lindau-Eschikon Research Station for Plant Sciences in Switzerland. We will consider active 
classification of crop-weed distributions on a 20-plot $40\times100$m sugarbeet field.

\subsubsection*{Acknowledgments.} This work was funded by the European
Community's Horizon 2020 programme under grant agreement no 644227-Flourish and
from the Swiss State Secretariat for Education, Research and Innovation (SERI)
under contract number 15.0029. We would like to thank the ETH Crop Science
Group for providing the testing facilities.

\bibliographystyle{splncsnat}
\footnotesize
\bibliography{references/IEEEabrv,references/strings-supershort,references/2016-iser-popovic}
%\bibliography{references/IEEEabrv,references/strings-short,references/library}

\end{document}

%% file: common_acronyms.tex
\begin{acronym}

\acro{2D}{two-dimensional}

\acro{3D}{three-dimensional}

\acro{AHRS}{attitude and heading reference system}

\acro{AUV}{autonomous underwater vehicle}

\acro{CPP}{Chinese Postman Problem}

\acro{DoF}{degree of freedom}
\acrodefplural{DoF}[DoFs]{degrees of freedom}

\acro{DVL}{Doppler velocity log}

\acro{FSM}{finite state machine}

\acro{IMU}{inertial measurement unit}

\acro{LBL}{Long Baseline}

\acro{MCM}{mine countermeasures}

\acro{MDP}{Markov decision process}
\acrodefplural{MDP}[MDPs]{Markov decision processes}

\acro{POMDP}{partially observable Markov decision process}
\acrodefplural{POMDP}[POMDPs]{partially observable Markov decision processes}

\acro{PRM}{Probabilistic Roadmap}
\acrodefplural{PRM}[PRM]{Probabilistic Roadmaps}

\acro{ROI}{region of interest}
\acrodefplural{ROI}[ROIs]{regions of interest}

\acro{ROS}{Robot Operating System}

\acro{ROV}{remotely operated vehicle}

\acro{RRT}{Rapidly-exploring Random Tree}
\acrodefplural{RRT}[RRTs]{Rapidly-exploring Random Trees}

\acro{SLAM}{simultaneous localization and mapping}

\acro{SSE}{sum of squared errors}

\acro{STOMP}{Stochastic Trajectory Optimization for Motion Planning}

\acro{TRN}{Terrain-Relative Navigation}

\acro{UAV}{unmanned aerial vehicle}

\acro{USBL}{Ultra-Short Baseline}

\acro{IPP}{informative path planning}

\acro{FoV}{field of view}
\acrodefplural{FoV}[FoVs]{fields of view}

\end{acronym}